\documentclass[runningheads]{llncs}

\usepackage{eccv}
\usepackage{eccvabbrv}

\usepackage{eccvabbrv}

\usepackage[table,dvipsnames]{xcolor} 

\usepackage[utf8]{inputenc}
\usepackage{graphicx}   
\usepackage{booktabs}   
\usepackage{multirow}   
\usepackage{colortbl}   
\usepackage{geometry}   
\usepackage{amssymb}    
\usepackage{amsmath}    
\usepackage{amsmath}
\usepackage{graphicx}
\usepackage{subcaption} 
\usepackage{xcolor}    
\usepackage{calc}      
\DeclareMathOperator*{\argmin}{arg\,min}
\definecolor{headergray}{gray}{0.96}           
\definecolor{sectiongray}{RGB}{248, 248, 248}  
\definecolor{rowgray}{gray}{0.93}              
\definecolor{lotusblue}{RGB}{235, 245, 255}    

\definecolor{forestgreen}{RGB}{47, 158, 68}    
\definecolor{brickred}{RGB}{200, 30, 30}       
\definecolor{darkgreen}{RGB}{39, 174, 96}      

\newcommand{\deltametric}[3]{{\tiny\textcolor{#1}{(\ensuremath{#2} #3)}}}

\usepackage[accsupp]{axessibility}  

\usepackage[pagebackref,breaklinks,colorlinks,citecolor=eccvblue]{hyperref}

\usepackage{orcidlink}

\begin{document}

\title{ForensicZip: More Tokens are Better but Not Necessary in Forensic Vision-Language Models} 

\titlerunning{Abbreviated paper title}

 \author{
    Yingxin Lai\inst{1} \and
    Zitong Yu\inst{1}\thanks{Corresponding authors.} \and
    Jun Wang\inst{1}$^{\star}$ \and
    Linlin Shen\inst{2} \and
    Yong Xu\inst{3} \and
    Xiaochun Cao\inst{4}
}

\authorrunning{Y. Lai et al.}

\institute{
    Great Bay University \and
    Shenzhen University \and
    Harbin Institute of Technology \and
    School of Cyber Science and Technology, Sun Yat-sen University\\
    \email{\{yingxinlai2@\}@gmail.com}  
}

\maketitle

\begin{abstract}
 
 Multimodal Large Language Models (MLLMs) enable interpretable multimedia forensics by generating textual rationales for forgery detection. However, processing dense visual sequences incurs high computational costs, particularly for high-resolution images and videos. Visual token pruning is a practical acceleration strategy, yet existing methods are largely semantic-driven, retaining salient objects while discarding background regions where manipulation traces such as high-frequency anomalies and temporal jitters often reside. To address this issue, we introduce ForensicZip, a training-free framework that reformulates token compression from a forgery-driven perspective. ForensicZip models temporal token evolution as a Birth-Death Optimal Transport problem with a slack dummy node, quantifying physical discontinuities indicating transient generative artifacts. The forensic scoring further integrates transport-based novelty with high-frequency priors to separate forensic evidence from semantic content under large-ratio compression. Experiments on deepfake and AIGC benchmarks show that at 10\% token retention, ForensicZip achieves $2.97\times$ speedup and over 90\% FLOPs reduction while maintaining state-of-the-art detection performance.  Code is publicly available at \url{https://github.com/laiyingxin2/ForensicZip}.
\end{abstract}

\section{Introduction}
\label{sec:intro}

Recent advances in Multimodal Large Language Models (MLLMs)~\cite{llava1.5,llavanext,lin2023video-llava,Qwen2VL,bai2023qwen-vl,mobilevlmv1,wen2025efficient} have led to remarkable progress in building generalist systems for diverse visual understanding and reasoning tasks. In multimedia forensics, MLLMs are increasingly adopted as an explainable pathway for forgery detection. With large-scale visual instruction tuning, forensic MLLMs~\cite{xu2024fakeshield, zhou2025aigi, tan2025veritas, huang2025sida, guo2025rethinking} can not only predict authenticity but also produce fine-grained textual rationales, such as pointing out abnormal clues, blending boundaries, or inconsistent reflections~\cite{wen2025spot,chen2024x2}. However, practical deployment remains challenging. To capture subtle manipulation traces, these models must process high-resolution images or long video streams. This mechanism generates a massive number of visual tokens, making the prefilling stage a computational bottleneck due to the heavy computational cost cost of self-attention.

A growing line of work shows that visual tokens in Vision-Language Models (VLMs) and MLLMs are highly redundant, motivating training-free token pruning and merging as a practical inference acceleration strategy~\cite{han2024filter, yu2026visiontrim}. Representative methods perform early-layer pruning~\cite{chen2024image,wen2025token}, attention- or sparsity-guided token selection~\cite{zhang2024sparsevlm,shang2024llava}, or adaptive token reduction~\cite{shang2025llava, sun2025llava}. Most of these approaches rank tokens by semantic salience (e.g., cross-modal attention, vision-language similarity, or macro-diversity~\cite{alvar2025divprune, li2025balanced}) and achieve substantial compute savings with minor performance drops on standard benchmarks such as captioning~\cite{Fu2023MMEAC,yue2024mmmu} and VQA~\cite{zhou2025urbench,yu2019activitynet}.

Despite this progress, semantic-driven token reduction is often misaligned with the forgery-driven requirements of forensic detection. Key manipulation cues (such as high-frequency noise patterns, subtle blending seams, or transient temporal jitters) tend to be weakly correlated with object-level semantics. They are often distributed in visually flat regions that semantic criteria treat as redundant~\cite{zhong2024aim,wen2025token,yu2026visiontrim}. Meanwhile, recent analyses report that text-guided attention can exhibit cross-modal misalignment inside MLLM layers, making semantic scores unreliable under compression~\cite{xu2025rethinking, zhang2025beyond}. Consequently, current methods tend to over-retain macro-semantic objects while discarding low-saliency regions that still contain forensic evidence, leading to significant performance degradation at high pruning ratios, as illustrated in Fig.~\ref{fig:preliminary_obs}(a). These observations motivate three questions:
(1) Why do existing pruning methods suffer task-dependent degradation on  forgery detection?
(2) How do generative artifacts manifest dynamically in the visual token space, especially over time?
(3) How can we aggressively compress visual tokens while preserving subtle, non-salient forensic evidence?

\begin{figure*}[t]
\centering
\vspace{-1.2mm}
\includegraphics[width=0.7\linewidth]{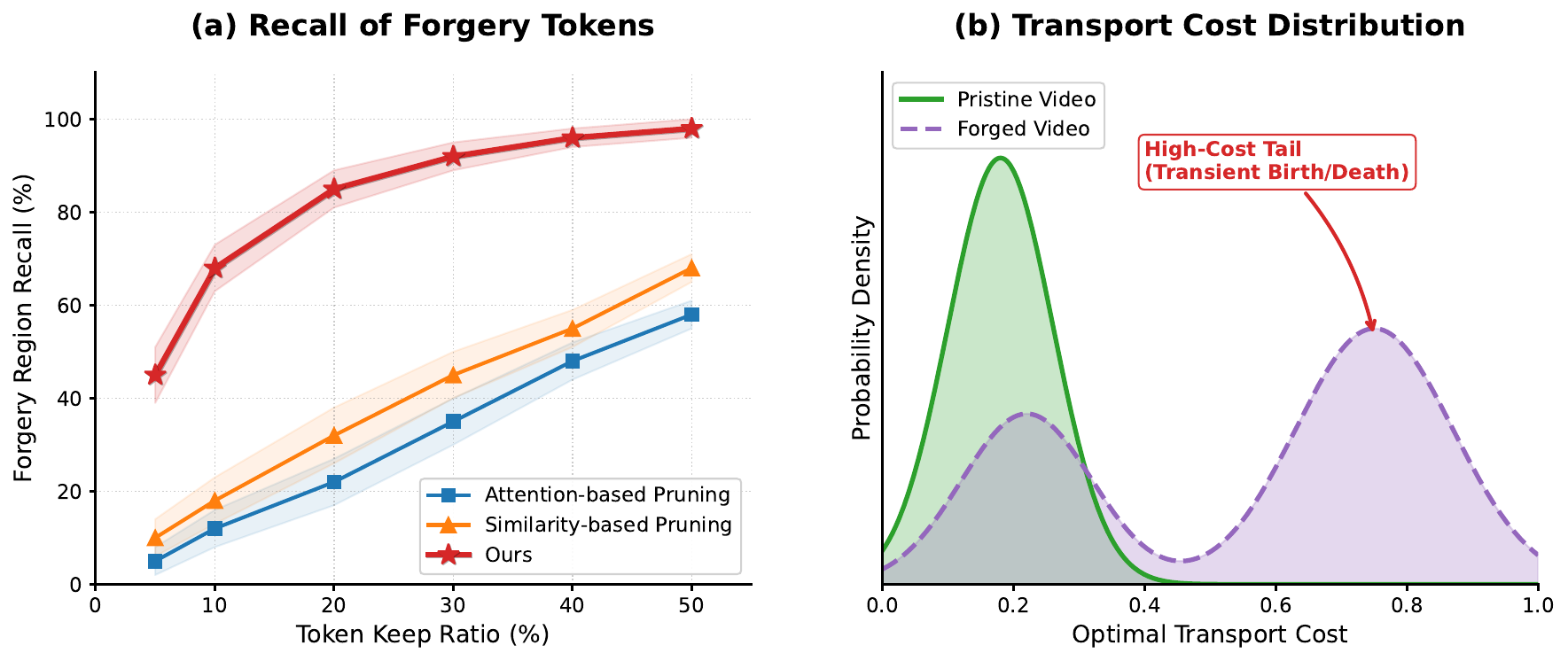}
 \caption{\textbf{Quantitative analysis of forensic-semantic misalignment.}
 \vspace{-0.4em}
(a) Recall of retained tokens against ground-truth forgery regions under varying compression ratios. Semantic-driven criteria show high sensitivity to compression ratios, while our transport-based novelty score maintains stable coverage of localized anomalies. Shaded regions denote $\pm 1$ standard deviation across evaluation samples.
(b) Distribution of temporal Optimal Transport (OT) costs. The distinct high-cost tail in forged videos indicates physical discontinuities (birth-death events), supporting the use of our augmented OT formulation.}
\label{fig:preliminary_obs}
\vspace{-2.3em}
\end{figure*}

In this work, we observe that many modern generation pipelines, despite strong per-frame realism, often violate physical continuity across frames. In the visual token space, such violations manifest as transient birth and death events of local textures and structures (Fig.~\ref{fig:preliminary_obs}b). Building on this insight, we propose ForensicZip, a training-free token compression framework that shifts the paradigm from semantic-driven pruning to forgery-driven sparsification. ForensicZip consists of two core components:
First, Transport Novelty Estimation (TNE) formulates temporal token evolution as an augmented entropic Optimal Transport problem. By introducing a dummy node, it explicitly translates physical incoherence into quantifiable birth and death costs.
Second, Forensic Scoring integrates these transport costs with frequency-domain priors and applies distance-penalized stabilization to separate forensic signals from semantic content. Finally, based on these stabilized scores, we perform a physical Top-$K$ selection, yielding genuine sequence-length reduction and end-to-end acceleration.
Our contributions are summarized as follows:
\begin{itemize}
    \item \textbf{Task-specific Analysis:} We identify a structural mismatch between semantic-driven token pruning and the forgery-driven nature of forgery detection, demonstrating that preserving global anchors and transient artifacts is crucial under aggressive token budgets.
    \item \textbf{ForensicZip Framework:} We propose a training-free token sparsification framework that leverages augmented Birth-Death Optimal Transport for forensic novelty estimation, together with a frequency-aware scoring strategy for robust evidence preservation.
    \item \textbf{Performance Validation:} ForensicZip seamlessly integrates into existing forensic MLLMs via physical pruning. Extensive experiments across deepfake and AIGC benchmarks demonstrate that at an aggressive 10\% token retention, our method achieves 2.97$\times$ speedup and reduces FLOPs by over 90\% while maintaining state-of-the-art detection performance.
\end{itemize}
\section{Related Work}
\label{sec:related}

\paragraph{Synthetic Image Detection via MLLMs.}
Deepfake and synthetic image detection have shifted from binary classification~\cite{cnn-detect,xception,detect-photoshop,shao2022detecting,heo2021deepfake} to interpretable forensics using Multimodal Large Language Models (MLLMs). Recent methods bridge visual anomalies with natural-language rationales via common-sense~\cite{zhang2024common} or pattern-aware reasoning~\cite{tan2025veritas}, and language-guided localization~\cite{guo2025language}. To enhance transparency, frameworks like FakeShield~\cite{xu2024fakeshield}, X2-DFD~\cite{chen2024x2}, and Spot the Fake~\cite{wen2025spot} generate fine-grained explanations paired with spatial masks. This paradigm extends to diverse domains, including face forensics~\cite{guo2025rethinking}, diffusion videos~\cite{song2024learning}, and social media (e.g., SIDA~\cite{huang2025sida}), culminating in comprehensive forensic assistants like AIGI-Holmes~\cite{zhou2025aigi} and LEGION~\cite{kang2025legion}. Despite these advances, these methods typically assume a sufficient computational budget to process dense visual tokens, neglecting the severe prefill latency bottleneck in high-resolution or long-video scenarios.
\vspace{-0.5em}
 \paragraph{Visual Token Compression for Efficient MLLM Inference.}
Visual token reduction mitigates redundancy and accelerates MLLM inference. One prominent line of research relies on semantic importance to rank tokens via early-layer~\cite{chen2024image,chen2024fastv} or text-to-vision attention~\cite{zhang2024sparsevlm}. To improve information density and coverage, recent approaches employ clustering~\cite{sun2025llava,zhong2024aim} or diversity-based selection~\cite{alvar2025divprune, li2025balanced} to prevent token over-concentration. For video and high-resolution scenarios, frameworks explicitly exploit spatio-temporal dynamics~\cite{wen2025token} or offer training-free plug-and-play pipelines~\cite{han2024filter, liu2025global} to reduce redundancy. Despite this progress, existing methods remain fundamentally semantic-driven, prioritizing macro-level visual understanding. In contrast, forensic cues are intrinsically anti-semantic localized artifacts like resampling patterns or temporal jitters frequently reside in regions deemed redundant by conventional metrics. Furthermore, recent analyses~\cite{fu2025exploring,  zhang2025beyond} reveal that semantic pruning under cross-modal misalignment can destroy the global spatial reference required for precise forgery localization. These limitations motivate our ForensicZip framework, which formulates token evolution as an optimal transport problem to preserve non-semantic forensic evidence while achieving physical sequence-length reduction.

 \section{ForensicZip}
\label{sec:forensiczip}

We first review the forensic VLM pipeline and analyse why visual
sequence length is the primary efficiency bottleneck
(Sec.~\ref{sec:fzip_pre}).  We then present two empirical observations
that reveal a structural mismatch between semantic-driven token pruning
and forgery detection (Sec.~\ref{sec:fzip_obs}).  Finally,
Sec.~\ref{sec:fzip_method} introduces our training-free scoring and
selection procedure, which retains a small token subset per frame based
on physical inconsistency rather than conceptual saliency.

\subsection{Preliminary}
\label{sec:fzip_pre}

\paragraph{Forensic VLM pipeline.}
A typical forensic VLM consists of a vision encoder, a multimodal
projector $g:\mathbb{R}^{D_v}\!\to\!\mathbb{R}^{D_p}$, and a language
model, where $D_v$ is the vision-encoder hidden dimension and $D_p$ is
the language embedding dimension.  Given a video with $T$ frames
$\{\mathbf{I}_t\}_{t=1}^{T}$, the vision encoder produces, for each
frame~$t$, one global token and $N$ patch tokens
$\mathbf{V}_t = \{\mathbf{v}_{t,1},\ldots,\mathbf{v}_{t,N}\}
\in\mathbb{R}^{N\times D_v}$, where
$\mathbf{v}_{t,i}\in\mathbb{R}^{D_v}$ is the $i$-th patch token.

\paragraph{Sequence-length complexity.}
For a transformer with $L_{\mathrm{layer}}$ layers, sequence length~$n$,
hidden size~$d$, and FFN intermediate size~$m$, the FLOPs are
approximated by
\begin{equation}
  \mathrm{FLOPs}
  = L_{\mathrm{layer}}\bigl(4nd^{2}+2n^{2}d+2ndm\bigr).
  \label{eq:fzip_flops}
\end{equation}
The quadratic self-attention term $2n^{2}d$ makes cost sensitive
to~$n$.  In multimodal prompting the total length decomposes as
\begin{equation}
  n = n_{\mathrm{sys}}+n_{\mathrm{vis}}+n_{\mathrm{txt}},
  \qquad
  n_{\mathrm{vis}} = T\,(N+1),
  \label{eq:fzip_len}
\end{equation}
where $n_{\mathrm{sys}}$, $n_{\mathrm{txt}}$, and $n_{\mathrm{vis}}$
denote system-prompt, user-text, and visual-token lengths, respectively.
Because $n_{\mathrm{vis}}$ dominates in high-resolution and video
settings, reducing visual tokens directly lowers prefill cost and memory
footprint~\cite{wen2025token,yu2026visiontrim,bolya2022tome,alvar2025divprune}.

\subsection{The Forensic-Semantic Discrepancy}
\label{sec:fzip_obs}

Before presenting our framework, we investigate how common token
reduction paradigms behave in the forensic domain.  Through systematic
analysis on the FakeVLM benchmark, we identify two misalignments that
together motivate a transport-based scoring mechanism with explicit
slack.

\paragraph{Obs.\,1: Inverse correlation between semantic saliency and
forensic evidence.}
Existing pruning strategies assume that semantically salient regions
carry the most task-critical information.  As shown in
Fig.~\ref{fig:preliminary_obs}(a), however, we observe a systematic
inverse correlation between cross-modal attention magnitude and spatial
overlap with ground-truth forgery masks.  MLLM attention heads converge
on high-level object centres, whereas manipulation traces such as
blending inconsistencies, resampling grids, and compression artefacts
reside predominantly in low-saliency backgrounds and object edges.
Purely semantic reduction therefore acts as a low-pass filter on forensic
information, preserving convincing content while removing the subtle
evidence required for detection.

\vspace{-1.0em}
\paragraph{Obs.\,2: Temporal discontinuity as Birth--Death events.}
We further examine token dynamics in the projected feature space across
adjacent frames.  In pristine video, tokens evolve smoothly: under
uniform marginals the optimal transport plan between consecutive frames
concentrates mass along the cost-matrix diagonal, reflecting continuous
spatial correspondence.  Generative models violate this continuity:
certain features appear without a plausible source (Birth) or vanish
without a matching target (Death).  As shown in
Fig.~\ref{fig:preliminary_obs}(b), conventional nearest-neighbour or
balanced matching forces these outliers into spurious correspondences.
Because the anomalous mass is then distributed across many low-cost
entries rather than concentrated in a single high-cost event, the forgery
signal is diluted below a practical detection threshold.  This dilution
effect motivates an unbalanced transport formulation that can explicitly
route unmatchable mass through a dedicated slack node.

\begin{figure}[t]
  \centering
  \includegraphics[width=0.8\linewidth]{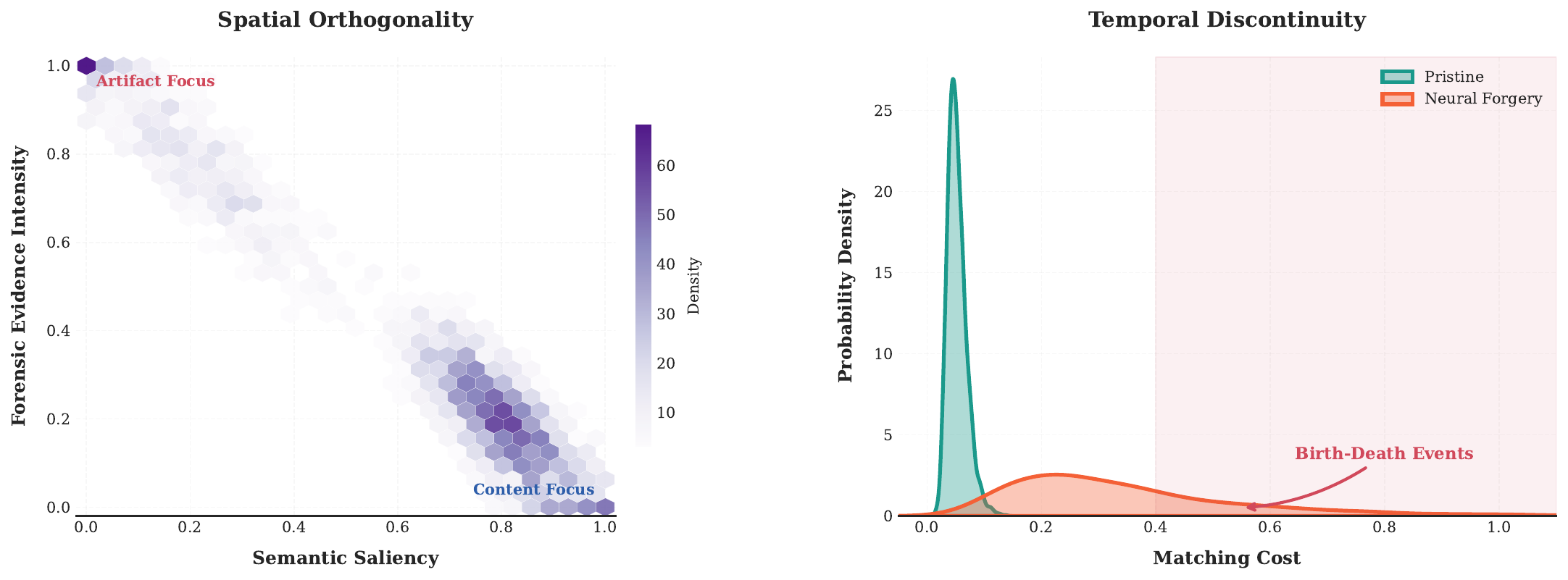}
  \caption{\textbf{Empirical motivation for ForensicZip.}
  \textbf{(a)~Inverse Correlation:} Density plot of semantic saliency
  (cross-modal attention) vs.\ forensic evidence (IoU with forgery masks)
  on FakeClue/LOKI.
  \textbf{(b)~Temporal Discontinuity:} KDE of inter-frame matching costs.}
  \label{fig:preliminary_obs}
  \vspace{-2em}
\end{figure}

\subsection{Methodology}
\label{sec:fzip_method}

\begin{figure*}[t]
  \centering
  \includegraphics[width=0.9\linewidth]{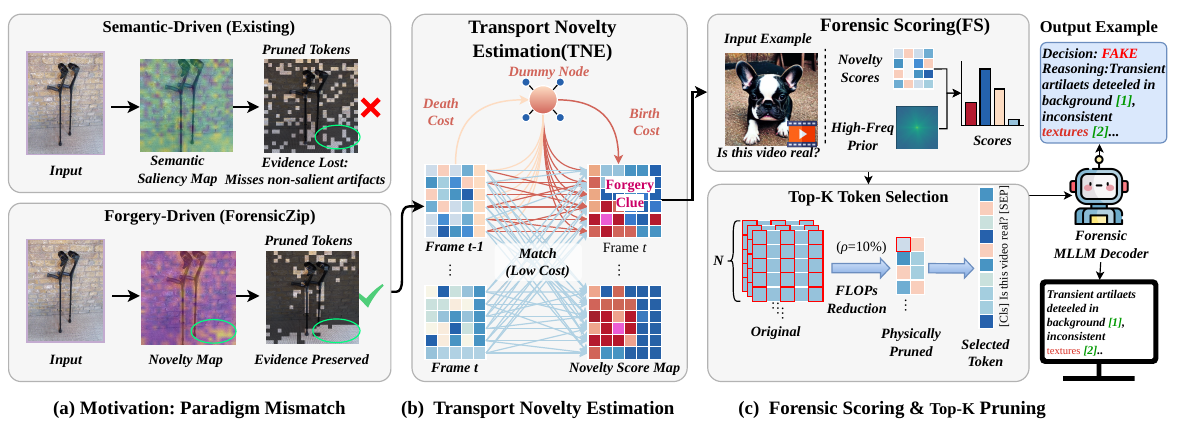}
  \vspace{-1.0em}
  \caption{\textbf{Overall framework of ForensicZip.} ForensicZip
  measures transport novelty across adjacent video frames and selectively
  preserves vision tokens with high physical inconsistency (i.e.,
  transient artifacts), thereby achieving plug-and-play forensic MLLM
  inference acceleration.}
  \label{fig:framework}
  \vspace{-2.0em}
\end{figure*}

Both observations point to a single principle: forensic evidence resides
where physical conservation laws are violated, not where semantic content
is concentrated.  Generation pipelines leave frequency-domain fingerprints
in the spatial domain (Obs.\,1) and break feature-level mass conservation
in the temporal domain (Obs.\,2).  We realise this principle through two
complementary stages (Fig.~\ref{fig:framework}).
\emph{Transport Novelty Estimation}~(TNE) detects temporal violations by
modelling inter-frame correspondence as an augmented optimal transport
problem with a slack node, so that unmatchable mass surfaces as
concentrated Birth--Death costs.  \emph{Forensic Scoring}~(FS) detects
spatial violations by modulating the temporal scores with a
high-frequency prior.  Their product implements a soft AND gate: only
tokens anomalous in both domains receive high scores.

\subsubsection{Transport Novelty Estimation (TNE)}
\label{sec:tne}

Forgery detection in the temporal domain reduces to conservation
violation detection: natural video preserves feature-level mass across
frames, whereas generative artefacts create or destroy local textures
abruptly.  A na\"ive cosine-similarity approach treats each token
independently, so anomalous residuals are absorbed by the single best
match.  Optimal transport improves on this by finding a global assignment
under marginal constraints, but standard balanced OT enforces strict mass
conservation, forcing every target token onto a real source and smearing
the anomaly across many small residuals.  Adding a dummy node closes this
gap: tokens that genuinely lack a predecessor can route their mass
through the dummy at a fixed penalty, surfacing as concentrated high-cost
events.

\paragraph{Birth--Death entropic optimal transport.}
Each patch token $\mathbf{v}_{t,i}\in\mathbb{R}^{D_v}$
($i=1,\ldots,N$) in frame~$t$ is projected and $\ell_2$-normalised:
\begin{equation}
  \hat{\mathbf{z}}_{t,i}
  = \frac{g(\mathbf{v}_{t,i})}
         {\lVert g(\mathbf{v}_{t,i})\rVert_2
          +\epsilon_{\mathrm{norm}}},
  \qquad
  \hat{\mathbf{z}}_{t,i}\in\mathbb{R}^{D_p},
  \label{eq:fzip_proj}
\end{equation}
where $\epsilon_{\mathrm{norm}}$ is a small constant for numerical
stability.  Normalisation maps all embeddings onto the unit hypersphere,
naturally yielding cosine distance as the ground cost.

For consecutive frames $(t{-}1,\,t)$ with $T\ge 2$, we build a pairwise
cost matrix $\mathbf{C}^{(t)}\in\mathbb{R}^{N\times N}$ whose entry
$C^{(t)}_{ij}
 =1-\hat{\mathbf{z}}_{t-1,\,i}^{\top}\,\hat{\mathbf{z}}_{t,\,j}$
measures the cosine distance between the $i$-th source token in
frame~$t{-}1$ and the $j$-th target token in frame~$t$.  To flag tokens
that appear without a source (\emph{Birth}) or vanish without a target
(\emph{Death}), we extend this matrix with a dummy row and column:
\begin{equation}
  \bar{\mathbf{C}}^{(t)}
  = \begin{bmatrix}
      \mathbf{C}^{(t)}
        & c_{\mathrm{death}}\,\mathbf{1}_{N} \\[4pt]
      c_{\mathrm{birth}}\,\mathbf{1}_{N}^{\top}
        & 0
    \end{bmatrix}
  \in\mathbb{R}^{(N+1)\times(N+1)},
  \label{eq:fzip_cost_aug}
\end{equation}
where $\mathbf{1}_{N}\in\mathbb{R}^{N}$ is the all-ones vector.  The
scalar penalties $c_{\mathrm{birth}}\ge 0$ and
$c_{\mathrm{death}}\ge 0$ set the cost of routing mass through the dummy
node: whenever the cheapest real assignment of a target token exceeds
$c_{\mathrm{birth}}$, the solver preferentially explains it as a newly
appeared artefact.  Instead of forcing a spurious pairing that dilutes
the anomaly, the dummy node absorbs the unmatchable mass and
concentrates the cost into a single interpretable entry.

We impose marginals
$\mathbf{a}
 =[\tfrac{1}{N},\ldots,\tfrac{1}{N},\,1]^{\top}
 \!\in\mathbb{R}^{N+1}$,
where the first $N$ entries assign uniform mass to real tokens and the
last entry provides unit slack to absorb all unmatchable mass if
necessary.  The entropic transport plan
$\bar{\mathbf{P}}^{(t)}\in\mathbb{R}^{(N+1)\times(N+1)}$ is obtained
via $I_{\mathrm{sk}}$ Sinkhorn iterations (set to 20 in all
experiments):
\begin{equation}
  \bar{\mathbf{P}}^{(t)}
  = \argmin_{\substack{
        \mathbf{P}\ge 0,\\[2pt]
        \mathbf{P}\,\mathbf{1}_{N+1}=\mathbf{a},\;
        \mathbf{P}^{\top}\mathbf{1}_{N+1}=\mathbf{a}
    }}\;
    \bigl\langle
      \mathbf{P},\,\bar{\mathbf{C}}^{(t)}
    \bigr\rangle_{F}
    +\varepsilon_{\mathrm{ot}}
     \sum_{i=1}^{N+1}\sum_{j=1}^{N+1}
     P_{ij}\bigl(\log P_{ij}-1\bigr),
  \label{eq:fzip_ot}
\end{equation}
where $\langle\cdot,\cdot\rangle_{F}$ denotes the Frobenius inner
product, $\varepsilon_{\mathrm{ot}}>0$ is the entropic regularisation
strength, and $\mathbf{1}_{N+1}\in\mathbb{R}^{N+1}$ is the
$(N{+}1)$-dimensional all-ones vector used in the marginal constraints.

\paragraph{Per-token score extraction.}
From the solved plan we extract two scores for each target token~$j$
($j=1,\ldots,N$) in frame~$t$, spanning the full range of temporal
anomaly from gradual drift to abrupt appearance.
The \emph{transport cost} aggregates the weighted matching cost from all
$N$ real source tokens in frame~$t{-}1$, excluding the dummy row:
\begin{equation}
  e^{(t)}_{j}
  = \sum_{i=1}^{N}
    \bar{P}^{(t)}_{ij}\,C^{(t)}_{ij}\,.
  \label{eq:fzip_ecost}
\end{equation}
A high $e^{(t)}_{j}$ indicates that token~$j$ deviates from every
predecessor yet was still partially matched, capturing distributed
anomalies.  The \emph{Birth evidence}
\begin{equation}
  b^{(t)}_{j}
  = \bar{P}^{(t)}_{N+1,\,j}
  \label{eq:fzip_birth}
\end{equation}
records the mass routed from the dummy source, directly quantifying the
absence of a plausible origin in frame~$t{-}1$ and capturing abrupt
appearance that admits no real correspondence.

For single-image inputs ($T=1$), the module falls back to a spatial
mode: we compute the frame-level mean embedding
$\boldsymbol{\mu}=\tfrac{1}{N}\sum_{j=1}^{N}\hat{\mathbf{z}}_{1,j}$
and define per-token novelty as
$s^{(1)}_{j}=\|\hat{\mathbf{z}}_{1,j}-\boldsymbol{\mu}\|_2$.

Transport cost measures feature-level displacement but is agnostic to
its cause: legitimate large-scale motion such as camera panning produces
elevated cost indistinguishable from a generative artefact on the basis
of temporal signal alone.  Resolving this ambiguity requires a signal
from a different physical domain.

\subsubsection{Forensic Scoring and Selection (FS)}
\label{sec:fs}

Generative artefacts and natural motion differ not in temporal
displacement magnitude but in spatial frequency signature.  Natural
motion displaces existing textures without altering their local spectral
statistics, whereas generation pipelines introduce characteristic
fingerprints: aliasing from resampling, spectral gaps from upsampling
networks, and gradient discontinuities at blending boundaries.  This
frequency-domain signal also addresses the inverse correlation identified
in Obs.\,1: forensic traces in semantically flat regions still carry
strong high-frequency responses and are therefore preserved.

\paragraph{High-frequency prior integration.}
We apply a fixed $3\times 3$ Laplacian operator $\mathcal{L}(\cdot)$ to
the input frame~$\mathbf{I}_t$ and average-pool the response to the
$N$-element patch grid, yielding a spatial proxy
$\mathbf{U}^{(t)}\in[0,1]^{N}$ where $U^{(t)}_{j}$ is the normalised
response at the $j$-th patch position.  The forensic score for token~$j$
in frame~$t$ is the product of a temporal anomaly term and a spatial
modulation term:
\begin{equation}
  s^{(t)}_{j}
  = \underbrace{\bigl(
      e^{(t)}_{j}
      +\lambda_{\mathrm{birth}}\,b^{(t)}_{j}
    \bigr)}_{\text{temporal anomaly}}
    \;\cdot\;
    \underbrace{\bigl(
      1+\eta_{\mathrm{forensic}}\,U^{(t)}_{j}
    \bigr)}_{\text{spatial modulation}},
  \label{eq:fzip_final_score}
\end{equation}
where $\lambda_{\mathrm{birth}}\ge 0$ balances Birth evidence against
transport cost, and $\eta_{\mathrm{forensic}}\ge 0$ controls the
amplification from high-frequency features.  The multiplicative form
implements a soft AND gate: a token must carry both temporal novelty and
spatial edge activity to score highly.  An additive alternative would
let a uniformly displaced background patch (high transport cost, zero
frequency response) dominate the ranking during camera panning, while a
static object boundary (strong edges, zero temporal novelty) would
collapse the score into a generic edge detector.  The product suppresses
both cases because zeroing either factor eliminates the score.

\paragraph{Token selection.}
Given a retention ratio $\rho\in(0,1]$, we select the top
$K=\lfloor\rho N\rfloor$ tokens from
$\{s^{(t)}_{j}\}_{j=1}^{N}$ for each frame~$t$, where $K$ denotes the
retained patch count per frame.  The global token is always kept.  The
compressed sequence of length $T(K{+}1)$ replaces the original
$T(N{+}1)$ tokens and is passed directly to the language model.

\subsubsection{Computational Complexity}
\label{sec:fz_complexity}

Each transformer layer incurs
$\mathcal{O}(n^{2}D_{m}+nD_{m}D_{\mathrm{ff}})$ cost, where $n$ is the
token count, $D_{m}$ is the language-model hidden dimension, and
$D_{\mathrm{ff}}$ is the FFN intermediate size.  Reducing the per-frame
visual sequence from $N{+}1$ to $K{+}1$ tokens across all
$L_{\mathrm{layer}}$ layers eliminates the dominant quadratic term by a
factor of $1-(K{+}1)^{2}/(N{+}1)^{2}$ per layer.  The only overhead is
the entropic OT solver on an $(N{+}1)\times(N{+}1)$ cost matrix:
\begin{equation}
  \mathrm{Cost}_{\mathrm{OT}}
  = \mathcal{O}\!\bigl(
      (T{-}1)\,I_{\mathrm{sk}}\,(N{+}1)^{2}
    \bigr),
  \label{eq:fz_ot_cost}
\end{equation}
where $I_{\mathrm{sk}}$ is the number of Sinkhorn iterations and
$T{-}1$ is the number of consecutive frame pairs.  Because this cost is
incurred once before the LLM forward pass and scales linearly with frame
count, it is negligible relative to the multi-layer transformer savings.
\textbf{A detailed theoretical analysis is provided in the supplementary
material.}

 \section{Experiments}
\label{sec:experiments}

 \begin{table*}[t]
\centering
\scriptsize
\renewcommand{\arraystretch}{1.28}
\setlength{\tabcolsep}{2.5pt}

\caption{\textbf{Quantitative Performance of ForensicZip on FakeVLM and FakeShield Backbones.}
``Vanilla'' denotes the full-token upper bound.
\textbf{Avg. Lat. (ms)} is end-to-end per-sample inference latency, \textbf{GPU Mem. (GB)} is peak GPU memory, and \textbf{Speedup} is w.r.t.\ Vanilla; efficiency metrics are averaged over all datasets in this table.
\textit{(\textcolor{red}{Red} text indicates performance collapse.)}}
\vspace{-4.2mm}
\label{tab:main_results}

\resizebox{\linewidth}{!}{
\begin{tabular}{l cccc cccc cccc}
\toprule[1.5pt]
\multirow{2.5}{*}{\textbf{Method}} &

\multicolumn{4}{c}{\textbf{Backbone: FakeVLM}} &
\multicolumn{4}{c}{\textbf{Backbone: FakeShield}} &
\multirow{2.5}{*}{\textbf{\shortstack{Avg.\\FLOPs(T)}}} &
\multirow{2.5}{*}{\textbf{\shortstack{Avg.\\Lat.(ms)}}} &
\multirow{2.5}{*}{\textbf{\shortstack{GPU\\Mem.(GB)}}} &
\multirow{2.5}{*}{\textbf{Speedup}} \\
\cmidrule(lr){2-5} \cmidrule(lr){6-9}
& \textbf{FakeClue} & \textbf{LOKI} & \textbf{DD-VQA} & \textbf{DMimage} & \textbf{PhotoShop} & \textbf{DeepFake} & \textbf{AIGC} & \textbf{DFFD} & & & & \\
\midrule

\rowcolor{sectiongray} \multicolumn{12}{c}{\textbf{\textit{Upper Bound (100\% Tokens)}}} \\
Vanilla & 98.60 & 84.30 & 93.20 & 94.00 & 94.43 & 98.00 & 93.00 & 98.00 & 3.102 & 4520.2 & 26.7 & 1.00$\times$ \\
\midrule

\rowcolor{sectiongray} \multicolumn{12}{c}{\textbf{\textit{Retain 50\% Tokens}} ($\downarrow$ 50\% visual tokens)} \\
FastV \tiny{(ECCV'24)} & 95.23 & 78.15 & 89.52 & 88.24 & 90.15 & 93.42 & 88.54 & 93.22 & 2.150 & 3228.6 & 26.6 & 1.40$\times$ \\
SparseVLM \tiny{(ICML'25)} & 96.14 & 79.53 & 90.45 & 89.66 & 91.24 & 94.15 & 89.23 & 94.44 & 2.155 & 3180.4 & 26.5 & 1.42$\times$ \\
LLaVA-Scissor \tiny{(NeurIPS'25)} & 97.22 & 81.84 & 91.81 & 91.13 & 92.83 & 95.84 & 91.42 & 95.85 & 2.140 & 3110.5 & 26.7 & 1.45$\times$ \\
VisionTrim \tiny{(ICLR'26)} & 97.87 & 82.25 & 92.92 & 92.34 & 93.54 & 96.92 & 92.15 & 96.52 & 2.135 & 3090.2 & 26.4 & 1.46$\times$ \\
\rowcolor{lotusblue} \textbf{ForensicZip (Ours)} & \textbf{98.50} & \textbf{84.35} & \textbf{93.25} & \textbf{94.05} & \textbf{94.48} & \textbf{98.05} & \textbf{93.05} & \textbf{98.04} & \textbf{1.852} & \textbf{2756.4} & \textbf{22.4} & \textbf{1.64$\times$} \\

\midrule
\rowcolor{sectiongray} \multicolumn{12}{c}{\textbf{\textit{Retain 25\% Tokens}} ($\downarrow$ 75\% visual tokens)} \\ 
FastV \tiny{(ECCV'24)} & 82.34 & 65.55 & 76.42 & 75.13 & 78.25 & 81.52 & 75.43 & 80.84 & 1.420 & 2510.5 & 26.5 & 1.80$\times$ \\
SparseVLM \tiny{(ICML'25)} & 85.63 & 68.22 & 79.65 & 78.54 & 81.42 & 83.15 & 78.56 & 83.22 & 1.415 & 2480.8 & 26.4 & 1.82$\times$ \\
LLaVA-Scissor \tiny{(NeurIPS'25)} & 91.45 & 75.53 & 86.24 & 85.42 & 86.54 & 90.45 & 85.13 & 90.92 & 1.390 & 2420.2 & 26.6 & 1.86$\times$ \\
VisionTrim \tiny{(ICLR'26)} & 94.83 & 79.15 & 89.56 & 89.24 & 90.25 & 94.14 & 89.55 & 94.83 & 1.385 & 2390.6 & 26.3 & 1.89$\times$ \\
\rowcolor{lotusblue} \textbf{ForensicZip (Ours)} & \textbf{98.46} & \textbf{83.93} & \textbf{93.12} & \textbf{93.88} & \textbf{94.32} & \textbf{97.88} & \textbf{92.85} & \textbf{97.82} & \textbf{0.985} & \textbf{2050.5} & \textbf{19.8} & \textbf{2.20$\times$} \\

\midrule
\rowcolor{sectiongray} \multicolumn{12}{c}{\textbf{\textit{Retain 10\% Tokens}} ($\downarrow$ 90\% visual tokens)} \\ 
FastV \tiny{(ECCV'24)} & \textcolor{brickred}{60.54} & \textcolor{brickred}{45.13} & \textcolor{brickred}{56.42} & \textcolor{brickred}{55.25} & \textcolor{brickred}{55.93} & \textcolor{brickred}{58.54} & \textcolor{brickred}{53.22} & \textcolor{brickred}{57.83} & 0.850 & 1950.2 & 26.4 & 2.31$\times$ \\
SparseVLM \tiny{(ICML'25)} & \textcolor{brickred}{65.82} & \textcolor{brickred}{48.45} & \textcolor{brickred}{60.14} & \textcolor{brickred}{59.53} & \textcolor{brickred}{60.25} & \textcolor{brickred}{63.83} & \textcolor{brickred}{58.14} & \textcolor{brickred}{61.52} & 0.845 & 1920.5 & 26.3 & 2.35$\times$ \\
LLaVA-Scissor \tiny{(NeurIPS'25)} & 78.25 & 62.53 & 72.56 & 71.24 & 72.54 & 76.12 & 71.43 & 75.85 & 0.820 & 1880.8 & 26.5 & 2.40$\times$ \\
VisionTrim \tiny{(ICLR'26)} & 84.53 & 70.84 & 80.25 & 78.56 & 79.14 & 82.55 & 76.82 & 82.24 & 0.815 & 1850.4 & 26.1 & 2.44$\times$ \\
\rowcolor{lotusblue} \textbf{ForensicZip (Ours)} & \textbf{97.74} & \textbf{82.15} & \textbf{90.58} & \textbf{91.23} & \textbf{92.15} & \textbf{95.88} & \textbf{90.85} & \textbf{95.62} & \textbf{0.452} & \textbf{1520.3} & \textbf{18.5} & \textbf{2.97$\times$} \\

\bottomrule[1.5pt]
\end{tabular}
}
\vspace{-4.6mm}
\end{table*}

\begin{table*}[t]
\centering
\scriptsize
\renewcommand{\arraystretch}{1.35} 
\setlength{\tabcolsep}{2.5pt}

\caption{\textbf{Efficiency comparisons on the FakeClue dataset.} LLM Generation Latency: time for LLM-only response generation; Model Generation Latency: time for the full model to generate response; Total Latency: total time to complete the FakeClue benchmark; GPU Peak Memory: maximum VRAM usage during inference; and Throughput: number of samples processed per second. Comparisons are conducted at \textbf{25\% token retention}.}
\vspace{-4.2mm} 
\label{tab:detailed_breakdown_fakeclue}

\resizebox{0.9\linewidth}{!}{
\begin{tabular}{l cccccc}
\toprule[1.5pt]
\multirow{2}{*}{\textbf{Methods}} & 
\textbf{LLM Generation}$\downarrow$ & 
\textbf{Model Generation}$\downarrow$ & 
\textbf{Total}$\downarrow$ & 
\textbf{GPU Peak}$\downarrow$ & 
\textbf{Throughput}$\uparrow$ & 
\textbf{Performance}$\uparrow$ \\

 & \textbf{Latency (s)} & \textbf{Latency (s)} & \textbf{Latency (min:sec)} & \textbf{Memory (GB)} & \textbf{(samples/s)} & \textbf{Acc (\%)} \\
\midrule

\rowcolor{sectiongray} 
\textit{FakeVLM-7B (Vanilla)} & 1472.5 & 1478.6 & 67:59 & 26.7 & 1.23 & 98.60 \\
\midrule

\multicolumn{7}{c}{\textit{\textbf{Retention Ratio = 25\%}}} \\

\textbf{FastV} \tiny{[ECCV'24]} & 
885.2 \deltametric{forestgreen}{\downarrow}{39.9\%} & 
892.4 \deltametric{forestgreen}{\downarrow}{39.6\%} & 
41:30 \deltametric{forestgreen}{\downarrow}{38.9\%} & 
26.5 \deltametric{forestgreen}{\downarrow}{0.7\%} & 
2.00 \deltametric{forestgreen}{\uparrow}{1.63$\times$} & 
82.34 \deltametric{brickred}{\downarrow}{16.2} \\

\textbf{SparseVLM} \tiny{[ICML'25]} & 
870.5 \deltametric{forestgreen}{\downarrow}{40.9\%} & 
878.2 \deltametric{forestgreen}{\downarrow}{40.6\%} & 
40:55 \deltametric{forestgreen}{\downarrow}{39.8\%} & 
26.6 \deltametric{forestgreen}{\downarrow}{0.4\%} & 
2.03 \deltametric{forestgreen}{\uparrow}{1.65$\times$} & 
85.63 \deltametric{brickred}{\downarrow}{12.9} \\

\textbf{LLaVA-Scissor} \tiny{[NeurIPS'25]} & 
852.1 \deltametric{forestgreen}{\downarrow}{42.1\%} & 
860.5 \deltametric{forestgreen}{\downarrow}{41.8\%} & 
40:10 \deltametric{forestgreen}{\downarrow}{40.9\%} & 
26.5 \deltametric{forestgreen}{\downarrow}{0.7\%} & 
2.07 \deltametric{forestgreen}{\uparrow}{1.68$\times$} & 
91.45 \deltametric{brickred}{\downarrow}{7.1} \\

\textbf{VisionTrim} \tiny{[ICLR'26]} & 
835.4 \deltametric{forestgreen}{\downarrow}{43.3\%} & 
842.1 \deltametric{forestgreen}{\downarrow}{43.0\%} & 
39:45 \deltametric{forestgreen}{\downarrow}{41.5\%} & 
26.4 \deltametric{forestgreen}{\downarrow}{1.1\%} & 
2.10 \deltametric{forestgreen}{\uparrow}{1.71$\times$} & 
94.83 \deltametric{brickred}{\downarrow}{3.7} \\

\rowcolor{lotusblue} 
\textbf{ForensicZip (Ours)} & 
\textbf{660.5} \deltametric{forestgreen}{\downarrow}{55.1\%} & 
\textbf{666.8} \deltametric{forestgreen}{\downarrow}{54.9\%} & 
\textbf{30:55} \deltametric{forestgreen}{\downarrow}{54.5\%} & 
\textbf{19.8} \deltametric{forestgreen}{\downarrow}{25.8\%} & 
\textbf{2.70} \deltametric{forestgreen}{\uparrow}{2.20$\times$} & 
\textbf{98.46} \deltametric{brickred}{\downarrow}{0.1} \\

\bottomrule[1.5pt]
\end{tabular}
}
\vspace{-4.2mm}
\end{table*}

\begin{table*}[t]
\centering
\small
\renewcommand{\arraystretch}{1.3}
\setlength{\tabcolsep}{3.5pt} 

\caption{\textbf{Comprehensive efficiency and performance results on the POPE~\cite{Li-hallucination-2023} benchmark.} Comparison of generation latency, end-to-end time, peak GPU memory, throughput, and accuracy on a single NVIDIA A100 GPU.}
\vspace{-4.2mm} 
\label{tab:comprehensive_efficiency}

\resizebox{\linewidth}{!}{
\begin{tabular}{l l c c c c c c c}
\toprule[1.5pt]
\textbf{Backbone} & \textbf{Method} & \textbf{\#Tokens} & \textbf{\shortstack{LLM Gen.\\Lat. (s)}} $\downarrow$ & \textbf{\shortstack{Total Lat.\\(min:sec)}} $\downarrow$ & \textbf{\shortstack{Peak Mem.\\(GB)}} $\downarrow$ & \textbf{\shortstack{Throughput\\(samples/s)}} $\uparrow$ & \textbf{\shortstack{FLOPs\\(T)}} $\downarrow$ & \textbf{Acc.} $\uparrow$ \\
\midrule

\multirow{4}{*}{\textbf{LLaVA-1.5-7B}} 
 & Vanilla & 576 & 618.0 & 21:43 (1.00$\times$) & 18.5 & 0.64 & 3.8 & 85.9 \\
 & + SparseVLM \tiny{(ICML'25)} & 64 & 512.4 & 18:24 (1.18$\times$) & 20.2 & 0.76 & 1.3 & 75.1 \\
 & + VisionTrim \tiny{(ICLR'26)} & 64 & 448.5 & 16:10 (1.34$\times$) & 19.1 & 0.86 & 1.0 & 84.5 \\
 \rowcolor{lotusblue}
 & + \textbf{ForensicZip (Ours)} & \textbf{64} & \textbf{412.0} & \textbf{14:52 (1.46$\times$)} & \textbf{16.8} & \textbf{0.94} & \textbf{0.8} & \textbf{86.1} \\

\midrule

\multirow{4}{*}{\textbf{LLaVA-NeXT-7B}} 
 & Vanilla & 2880 & 1008.4 & 38:04 (1.00$\times$) & 46.2 & 0.35 & 12.6 & 86.3 \\
 & + SparseVLM \tiny{(ICML'25)} & 320 & 846.2 & 31:43 (1.20$\times$) & 48.5 & 0.42 & 2.5 & 78.5 \\
 & + VisionTrim \tiny{(ICLR'26)} & 320 & 625.4 & 21:50 (1.74$\times$) & 42.0 & 0.61 & 1.8 & 83.5 \\
 \rowcolor{lotusblue}
 & + \textbf{ForensicZip (Ours)} & \textbf{320} & \textbf{520.8} & \textbf{18:28 (2.06$\times$)} & \textbf{36.5} & \textbf{0.72} & \textbf{1.2} & \textbf{85.8} \\

\bottomrule[1.5pt]
\end{tabular}
}
\vspace{-5.4mm} 
\end{table*}

 \begin{table*}[t]
\centering
\vspace{-2mm} 
\scriptsize 
\renewcommand{\arraystretch}{1.3} 

\begin{minipage}{\linewidth}
\centering
\setlength{\tabcolsep}{4pt} 
\caption{\textbf{Ablation study of core components in ForensicZip.} Evaluated on the \textbf{SID-Set} benchmark using the \textbf{SIDA-7B} backbone at a \textbf{10\% token retention} ratio. We incrementally validate the effectiveness of Transport Novelty Estimation (TNE) and Forensic Scoring (FS). \textbf{Base} uses FastV semantic attention ranking with physical Top-$K$ pruning (global token always kept) under the same token budget.}
\vspace{-2mm} 
\label{tab:ablation_components}

\resizebox{0.75\linewidth}{!}{ 
\begin{tabular}{l ccc | cc | cc | cc}
\toprule[1.5pt]
\multirow{2.5}{*}{\textbf{Variant}} & \multicolumn{3}{c|}{\textbf{Components}} & \multicolumn{2}{c|}{\textbf{Real Images}} & \multicolumn{2}{c|}{\textbf{Tampered Images}} & \multicolumn{2}{c}{\textbf{Overall Performance}} \\ 
\cmidrule(lr){2-4} \cmidrule(lr){5-6} \cmidrule(lr){7-8} \cmidrule(lr){9-10}
 & \textbf{TNE} & \textbf{FS} & \textbf{HF} & \textbf{Acc} & \textbf{F1} & \textbf{Acc} & \textbf{F1} & \textbf{Acc} & \textbf{F1} \\ 
\midrule
(a) Base & & & & 55.23 & 52.45 & 54.23 & 51.56 & 57.20 & 54.95 \\ 
(b) + TNE  & \checkmark & & & 80.10 & 78.60 & 82.20 & 80.30 & 83.40 & 82.10 \\
(c) + FS & \checkmark & \checkmark & & 85.60 & 87.40 & 88.20 & 87.90 & 90.10 & 89.90 \\
\rowcolor{lotusblue} 
(d) \textbf{ForensicZip} & \checkmark & \checkmark & \checkmark & \textbf{88.15} & \textbf{90.10} & \textbf{90.35} & \textbf{90.15} & \textbf{92.50} & \textbf{92.52} \\
\bottomrule[1.5pt]
\end{tabular}
} 
\end{minipage}

\vspace{1.5mm} 

\begin{minipage}[t]{0.48\linewidth}
\centering
\setlength{\tabcolsep}{3pt} 
\caption{\textbf{Effect of Transport Formulations in TNE.} Evaluated on \textbf{SID-Set} with \textbf{SIDA-7B} (10\% retention). Explicitly modeling unmatchable tokens via dummy nodes is crucial for capturing generative artifacts.}
\vspace{-2mm} 
\label{tab:ablation_ot}

\resizebox{\linewidth}{!}{ 
\begin{tabular}{l | cc | cc | c}
\toprule[1.5pt]
\multirow{2.5}{*}{\textbf{Alignment Strategy}} & \multicolumn{2}{c|}{\textbf{Synthetic}} & \multicolumn{2}{c|}{\textbf{Overall}} & \multirow{2.5}{*}{\textbf{\shortstack{Rel.\\Gain}}} \\ 
\cmidrule(lr){2-3} \cmidrule(lr){4-5}
 & \textbf{Acc} & \textbf{F1} & \textbf{Acc} & \textbf{F1} & \\ 
\midrule
Hard Assignment & 86.40 & 86.00 & 79.54 & 79.10 & --- \\
Balanced OT & 92.90 & 92.70 & 86.10 & 85.80 & +8.25\% \\
Only-Birth Penalty & 95.80 & 95.60 & 89.20 & 88.90 & +12.15\% \\
\rowcolor{lotusblue} 
\textbf{Birth-Death OT} & \textbf{97.95} & \textbf{97.85} & \textbf{92.50} & \textbf{92.52} & \textbf{+16.31\%} \\
\bottomrule[1.5pt]
\end{tabular}
} 
\end{minipage}
\hfill
\begin{minipage}[t]{0.48\linewidth}
\centering
\setlength{\tabcolsep}{4pt} 
\caption{\textbf{Effect of Spatial Operators in FS.} Evaluated on \textbf{FakeClue} using \textbf{FakeVLM-7B} (10\% retention). Laplacian ($\mathcal{L}$) yields the best preservation of localized noise by providing isotropic high-frequency priors.}
\vspace{-2mm} 
\label{tab:ablation_spatial}

\resizebox{0.9\linewidth}{!}{ 
\begin{tabular}{l | c | c}
\toprule[1.5pt]
\multirow{2.5}{*}{\textbf{Spatial Operator $\mathcal{L}(\cdot)$}} & \multirow{2.5}{*}{\textbf{Accuracy (\%)}} & \multirow{2.5}{*}{\textbf{\shortstack{Rel.\\Gain}}} \\ 
  & & \\ 
\midrule
None ($\eta_{\mathrm{forensic}} = 0$) & 93.45 & --- \\
Patch Variance (RGB) & 94.82 & +1.47\% \\
Sobel (Directional) & 96.15 & +2.89\% \\
\rowcolor{lotusblue} 
\textbf{Laplacian (Isotropic, Ours)} & \textbf{97.74} & \textbf{+4.59\%} \\
\bottomrule[1.5pt]
\end{tabular}
} 
\end{minipage}

\vspace{-1.7em} 
\end{table*}

\begin{figure*}[t]
  \centering
  \includegraphics[width=\linewidth]{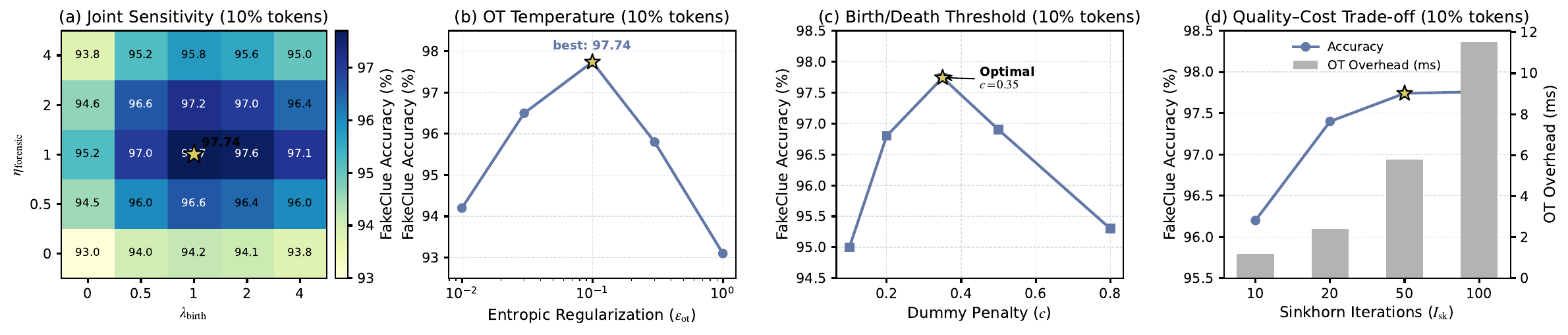}
 \caption{\textbf{Ablation Study.} Impact of (a) joint weights $\lambda/\eta$, (b) regularization $\varepsilon_{\mathrm{ot}}$, (c) penalty $c$, and (d) iterations on FakeClue accuracy.}
 \vspace{-1.8em}
  \label{fig:hyperparams}
\end{figure*}

  


\noindent \textbf{Benchmarks.}
To evaluate ForensicZip across diverse forgery types, we utilize three dataset categories: 
(1) \textbf{Universal \& AIGC Detection:} FakeVLM (FakeClue, LOKI~\cite{ye2024loki}), DMimage~\cite{corvi2023detection}, FakeShield-AIGC~\cite{xu2024fakeshield}, and SIDA~\cite{huang2024sida} for synthetic artifacts and social media imagery; 
(2) \textbf{DeepFake Forensics:} DD-VQA~\cite{zhang2024common} for artifact explanation, alongside FakeShield-DeepFake and DFFD~\cite{dang2020detectiondffd} for facial manipulation; and 
(3) \textbf{Traditional Manipulation:} FakeShield-PhotoShop (aggregating CASIA~\cite{dong2013casia}, etc.) for splicing and copy-move forgeries, and the commonly used POPE~\cite{Li-hallucination-2023} benchmark.

\noindent \textbf{Baselines.}
We compare ForensicZip against state-of-the-art training-free token pruning methods and full-token expert models:
\textbf{1) Training-free Token Pruning Methods:} We benchmark against leading inference acceleration strategies, including \textbf{FastV}~\cite{chen2024image}, \textbf{SparseVLM}~\cite{zhang2024sparsevlm}, and recent SOTAs such as \textbf{VisionTrim}~\cite{yu2026visiontrim} and \textbf{LLaVA-Scissor}~\cite{sun2025llava}.
\textbf{2) Forensic Expert Models (Vanilla):} We report the upper-bound performance using the full-token versions of state-of-the-art forensic MLLMs,  SIDA~\cite{huang2024sida}, FakeVLM~\cite{wen2025spot} and FakeShield~\cite{xu2024fakeshield} to demonstrate that ForensicZip retains expert-level capability under extreme compression.

 \noindent \textbf{Implementation Details.}
We adopt LLaVA-OneVision-7B~\cite{li2024llava-onevision} , SIDA-7B/13B~\cite{huang2024sida}, FakeVLM~\cite{wen2025spot} and FakeShield~\cite{xu2024fakeshield} as our primary backbones. The Transport Novelty Estimation (TNE) utilizes entropic regularization $\varepsilon=0.1$, with birth and death costs set to $c_{\mathrm{birth}}=c_{\mathrm{death}}=0.35$. All experiments are conducted on NVIDIA A100 (80GB) GPUs.



 \noindent\textbf{Performance on Forensic Benchmarks.}
We evaluate ForensicZip against leading training-free pruning baselines across diverse forgery types. As summarized in Tables~\ref{tab:main_results} and~\ref{tab:sida_full_results}, existing semantic-driven methods (FastV, SparseVLM) maintain competitive accuracy at conservative retention ratios but suffer catastrophic performance collapse as the token budget decreases. Specifically, at an aggressive 10\% retention ratio, these baselines degrade to near-random performance on the FakeClue and SID-Set benchmarks. In contrast, ForensicZip establishes a superior efficiency-accuracy frontier, maintaining 97.74\% accuracy on FakeClue with only 10\% visual tokens and scaling effectively to larger SIDA-13B backbones. This stability corroborates our core hypothesis: forensic evidence typically resides in semantically sparse regions. By prioritizing physical inconsistency rather than conceptual saliency, our transport-based scoring successfully preserves critical manipulation traces that are systematically filtered out by attention-based metrics.

\noindent\textbf{Efficiency and Robustness Analysis.}
We assess efficiency through both deployment latency and downstream reasoning stability. Unlike masking-based techniques that reduce attention complexity but retain full sequence lengths in FFN layers, ForensicZip implements physical token elimination. Table~\ref{tab:detailed_breakdown_fakeclue} confirms this design advantage: peak GPU memory consumption drops from 26.7GB to 19.8GB on LLaVA-OV-7B, translating to a $2.20\times$ throughput improvement. Furthermore, evaluation on the POPE object hallucination benchmark (Table~\ref{tab:comprehensive_efficiency}) reveals an unexpected benefit: ForensicZip slightly improves the accuracy of the vanilla LLaVA-1.5 model (86.1 vs 85.9). We attribute this precision gain to the removal of redundant semantic noise, which prevents the MLLM from over-focusing on irrelevant background context, thereby providing a cleaner evidence set for reasoning.

\noindent\textbf{Ablation Study.}
We validate component efficacy on the SID-Set benchmark (Table~\ref{tab:ablation_components}). Base semantic pruning yields poor results (57.20\% accuracy); introducing Transport Novelty Estimation (TNE) captures temporal dynamics, significantly boosting performance to 83.40\%. Table~\ref{tab:ablation_ot} further investigates the transport formulation: standard balanced OT smears forensic signals by forcing spurious correspondences, whereas our augmented Birth-Death mechanism explicitly isolates unmatchable artifacts via dummy nodes, yielding a substantial +16.31\% relative gain. Finally, Table~\ref{tab:ablation_spatial} indicates that the isotropic Laplacian operator outperforms directional filters, suggesting that generative artifacts manifest as non-directional high-frequency anomalies best captured by rotation-invariant second-order derivatives.

\noindent\textbf{Hyperparameter Sensitivity.} Fig.~\ref{fig:hyperparams} analyzes the impact of joint weights ($\lambda/\eta$), transport penalty ($c$), regularization ($\varepsilon$), and solver iterations. The performance remains stable across broad ranges (a-c), confirming the robustness of our dual-domain scoring.

\section{Conclusion}
\vspace{-1.3em}
 In this paper, we propose ForensicZip, a training-free and forgery-aware inference acceleration framework for forensic Vision-Language Models. Our insight for ForensicZip arises from our observation that forensic evidence is often spatially orthogonal to semantic saliency and manifests as temporal discontinuities in the token space. ForensicZip formulates temporal token evolution as an augmented birth death optimal transport problem to explicitly quantify transient artifacts, which results in significant computational cost reduction while preserving critical forensic cues even under aggressive compression.

\bibliographystyle{splncs04}
\bibliography{main}

@String(CVPR  = {IEEE Conf. Comput. Vis. Pattern Recog.})

@String(ICCV  = {Int. Conf. Comput. Vis.})

@String(ECCV  = {Eur. Conf. Comput. Vis.})

@String(ICML  = {Int. Conf. Mach. Learn.})

@String(ICLR  = {Int. Conf. Learn. Represent.})

@String(AAAI  = {AAAI})

@String(ICASSP=	{ICASSP})

@String(CVPR  = {CVPR})

@String(ICCV  = {ICCV})

@String(ECCV  = {ECCV})

@String(ICML  = {ICML})

@String(ICLR  = {ICLR})

@inproceedings{chen2024image,
  title={An image is worth 1/2 tokens after layer 2: Plug-and-play inference acceleration for large vision-language models},
  author={Chen, Liang and Zhao, Haozhe and Liu, Tianyu and Bai, Shuai and Lin, Junyang and Zhou, Chang and Chang, Baobao},
  booktitle={European Conference on Computer Vision},
  pages={19--35},
  year={2024},
  organization={Springer}
}

@article{bolya2022tome,
  title={Token merging: Your vit but faster},
  author={Bolya, Daniel and Fu, Cheng-Yang and Dai, Xiaoliang and Zhang, Peizhao and Feichtenhofer, Christoph and Hoffman, Judy},
  journal={arXiv preprint arXiv:2210.09461},
  year={2022}
}

@article{zhong2024aim,
  title={Aim: Adaptive inference of multi-modal llms via token merging and pruning},
  author={Zhong, Yiwu and Liu, Zhuoming and Li, Yin and Wang, Liwei},
  journal={arXiv preprint arXiv:2412.03248},
  year={2024}
}

@inproceedings{zhang2024sparsevlm,
  title={SparseVLM: Visual token sparsification for efficient vision-language model inference},
  author={Zhang, Yuan and Fan, Chun-Kai and Ma, Junpeng and Zheng, Wenzhao and Huang, Tao and Cheng, Kuan and Gudovskiy, Denis and Okuno, Tomoyuki and Nakata, Yohei and Keutzer, Kurt and others},
  booktitle={ICML},
  year={2025},
  organization={PMLR}
}

@inproceedings{chen2024fastv,
  title={An image is worth 1/2 tokens after layer 2: Plug-and-play inference acceleration for large vision-language models},
  author={Chen, Liang and Zhao, Haozhe and Liu, Tianyu and Bai, Shuai and Lin, Junyang and Zhou, Chang and Chang, Baobao},
  booktitle={ECCV},
  pages={19--35},
  year={2024},
  organization={Springer}
}

@article{li2024llava-onevision,
  title={Llava-onevision: Easy visual task transfer},
  author={Li, Bo and Zhang, Yuanhan and Guo, Dong and Zhang, Renrui and Li, Feng and Zhang, Hao and Zhang, Kaichen and Zhang, Peiyuan and Li, Yanwei and Liu, Ziwei and others},
  journal={arXiv preprint arXiv:2408.03326},
  year={2024}
}

@article{lin2023video-llava,
  title={Video-llava: Learning united visual representation by alignment before projection},
  author={Lin, Bin and Ye, Yang and Zhu, Bin and Cui, Jiaxi and Ning, Munan and Jin, Peng and Yuan, Li},
  journal={arXiv preprint arXiv:2311.10122},
  year={2023}
}

@article{bai2023qwen-vl,
  title={Qwen-vl: A frontier large vision-language model with versatile abilities},
  author={Bai, Jinze and Bai, Shuai and Yang, Shusheng and Wang, Shijie and Tan, Sinan and Wang, Peng and Lin, Junyang and Zhou, Chang and Zhou, Jingren},
  journal={arXiv preprint arXiv:2308.12966},
  year={2023}
}

@inproceedings{llava1.5,
  title={Improved baselines with visual instruction tuning},
  author={Liu, Haotian and Li, Chunyuan and Li, Yuheng and Lee, Yong Jae},
  booktitle={CVPR},
  pages={26296--26306},
  year={2024}
}

@misc{llavanext,
    title={LLaVA-NeXT: Improved reasoning, OCR, and world knowledge},
    url={https://llava-vl.github.io/blog/2024-01-30-llava-next/},
    author={Liu, Haotian and Li, Chunyuan and Li, Yuheng and Li, Bo and Zhang, Yuanhan and Shen, Sheng and Lee, Yong Jae},
    year={2024}
}

@article{Fu2023MMEAC,
  title={MME: A Comprehensive Evaluation Benchmark for Multimodal Large Language Models},
  author={Chaoyou Fu and Peixian Chen and Yunhang Shen and Yulei Qin and Mengdan Zhang and Xu Lin and Zhenyu Qiu and Wei Lin and Jinrui Yang and Xiawu Zheng and Ke Li and Xing Sun and Rongrong Ji},
  journal={arXiv preprint arXiv:2306.13394},
  year={2023}
}

@inproceedings{yu2019activitynet,
  title={Activitynet-qa: A dataset for understanding complex web videos via question answering},
  author={Yu, Zhou and Xu, Dejing and Yu, Jun and Yu, Ting and Zhao, Zhou and Zhuang, Yueting and Tao, Dacheng},
  booktitle={AAAI},
  volume={33},
  number={01},
  pages={9127--9134},
  year={2019}
}

@article{shang2024llava,
  title={Llava-prumerge: Adaptive token reduction for efficient large multimodal models},
  author={Shang, Yuzhang and Cai, Mu and Xu, Bingxin and Lee, Yong Jae and Yan, Yan},
  journal={arXiv preprint arXiv:2403.15388},
  year={2024}
}

@article{mobilevlmv1,
  title={Mobilevlm: A fast, reproducible and strong vision language assistant for mobile devices},
  author={Chu, Xiangxiang and Qiao, Limeng and Lin, Xinyang and Xu, Shuang and Yang, Yang and Hu, Yiming and Wei, Fei and Zhang, Xinyu and Zhang, Bo and Wei, Xiaolin and others},
  journal={arXiv:2312.16886},
  year={2023}
}

@inproceedings{zhang2024common,
  title={Common sense reasoning for deepfake detection},
  author={Zhang, Yue and Colman, Ben and Guo, Xiao and Shahriyari, Ali and Bharaj, Gaurav},
  booktitle={European Conference on Computer Vision},
  pages={399--415},
  year={2024},
  organization={Springer}
}

@inproceedings{ye2024loki,
  title={Loki: A comprehensive synthetic data detection benchmark using large multimodal models},
  author={Ye, Junyan and Zhou, Baichuan and Huang, Zilong and Zhang, Junan and Bai, Tianyi and Kang, Hengrui and He, Jun and Lin, Honglin and Wang, Zihao and Wu, Tong and others},
  booktitle={ICLR},
  year={2025},
}

@article{xu2024fakeshield,
  title={Fakeshield: Explainable image forgery detection and localization via multi-modal large language models},
  author={Xu, Zhipei and Zhang, Xuanyu and Li, Runyi and Tang, Zecheng and Huang, Qing and Zhang, Jian},
  journal={arXiv preprint arXiv:2410.02761},
  year={2024}
}

@article{huang2024sida,
  title={SIDA: Social Media Image Deepfake Detection, Localization and Explanation with Large Multimodal Model},
  author={Huang, Zhenglin and Hu, Jinwei and Li, Xiangtai and He, Yiwei and Zhao, Xingyu and Peng, Bei and Wu, Baoyuan and Huang, Xiaowei and Cheng, Guangliang},
  journal={arXiv preprint arXiv:2412.04292},
  year={2024}
}

@inproceedings{yue2024mmmu,
  title={Mmmu: A massive multi-discipline multimodal understanding and reasoning benchmark for expert agi},
  author={Yue, Xiang and Ni, Yuansheng and Zhang, Kai and Zheng, Tianyu and Liu, Ruoqi and Zhang, Ge and Stevens, Samuel and Jiang, Dongfu and Ren, Weiming and Sun, Yuxuan and others},
  booktitle={Proceedings of the IEEE/CVF Conference on Computer Vision and Pattern Recognition},
  pages={9556--9567},
  year={2024}
}

@article{Qwen2VL,
  title={Qwen2-VL: Enhancing Vision-Language Model's Perception of the World at Any Resolution},
  author={Wang, Peng and Bai, Shuai and Tan, Sinan and Wang, Shijie and Fan, Zhihao and Bai, Jinze and Chen, Keqin and Liu, Xuejing and Wang, Jialin and Ge, Wenbin and Fan, Yang and Dang, Kai and Du, Mengfei and Ren, Xuancheng and Men, Rui and Liu, Dayiheng and Zhou, Chang and Zhou, Jingren and Lin, Junyang},
  journal={arXiv preprint arXiv:2409.12191},
  year={2024}
}

@inproceedings{detect-photoshop,
  title={Detecting Photoshopped Faces by Scripting Photoshop},
  author={Wang, Sheng-Yu and Wang, Oliver and Owens, Andrew and Zhang, Richard and Efros, Alexei A},
  booktitle={ICCV},
  year={2019}
}

@inproceedings{cnn-detect,
  title={CNN-generated images are surprisingly easy to spot...for now},
  author={Wang, Sheng-Yu and Wang, Oliver and Zhang, Richard and Owens, Andrew and Efros, Alexei A},
  booktitle={CVPR},
  year={2020}
}

@inproceedings{xception,
  title = {Xception: Deep Learning with Depthwise Separable Convolutions},
  author = {Chollet, François},
  booktitle = {CVPR},
  year = {2017}
}

@inproceedings{fu2025exploring,
  title={Exploring Unbiased Deepfake Detection via Token-Level Shuffling and Mixing},
  author={Fu, Xinghe and Yan, Zhiyuan and Yao, Taiping and Chen, Shen and Li, Xi},
  booktitle={AAAI},
  year={2025}
}

@Misc{fs,
note = {\url{www.github.com/MarekKowalski/FaceSwap} Accessed 2021-04-24},
author = {FaceSwap.}
}

@article{heo2021deepfake,
  title={Deepfake detection scheme based on vision transformer and distillation},
  author={Heo, Young-Jin and Choi, Young-Ju and Lee, Young-Woon and Kim, Byung-Gyu},
  journal={arXiv preprint arXiv:2104.01353},
  year={2021}
}

@misc{deepfake,
  title = {DeepFakes},
  howpublished = {\url{https://github.com/iperov/DeepFaceLab}},
  note = {Accessed: 2020-05-10}
}

@inproceedings{corvi2023detection,
  title={On the detection of synthetic images generated by diffusion models},
  author={Corvi, Riccardo and Cozzolino, Davide and Zingarini, Giada and Poggi, Giovanni and Nagano, Koki and Verdoliva, Luisa},
  booktitle={ICASSP 2023-2023 IEEE International Conference on Acoustics, Speech and Signal Processing (ICASSP)},
  pages={1--5},
  year={2023},
  organization={IEEE}
}

@article{wen2025token,
  title={Token Pruning in Multimodal Large Language Models: Are We Solving the Right Problem?},
  author={Wen, Zichen and Gao, Yifeng and Li, Weijia and He, Conghui and Zhang, Linfeng},
  journal={arXiv preprint arXiv:2502.11501},
  year={2025}
}

@article{wen2025efficient,
  title={Efficient Multi-modal Large Language Models via Progressive Consistency Distillation},
  author={Wen, Zichen and Wang, Shaobo and Zhou, Yufa and Zhang, Junyuan and Zhang, Qintong and Gao, Yifeng and Chen, Zhaorun and Wang, Bin and Li, Weijia and He, Conghui and others},
  journal={arXiv preprint arXiv:2510.00515},
  year={2025}
}

@inproceedings{zhou2025urbench,
  title={Urbench: A comprehensive benchmark for evaluating large multimodal models in multi-view urban scenarios},
  author={Zhou, Baichuan and Yang, Haote and Chen, Dairong and Ye, Junyan and Bai, Tianyi and Yu, Jinhua and Zhang, Songyang and Lin, Dahua and He, Conghui and Li, Weijia},
  booktitle={Proceedings of the AAAI Conference on Artificial Intelligence},
  volume={39},
  number={10},
  pages={10707--10715},
  year={2025}
}

@String(CVPR= {IEEE Conf. Comput. Vis. Pattern Recog.})

@String(ICCV= {Int. Conf. Comput. Vis.})

@String(ECCV= {Eur. Conf. Comput. Vis.})

@String(ICLR = {Int. Conf. Learn. Represent.})

@String(AAAI = {AAAI})

@inproceedings{shao2022detecting,
  title={Detecting and recovering sequential deepfake manipulation},
  author={Shao, Rui and Wu, Tianxing and Liu, Ziwei},
  booktitle={European Conference on Computer Vision},
  pages={712--728},
  year={2022},
  organization={Springer}
}

@inproceedings{dong2013casia,
  title={Casia image tampering detection evaluation database},
  author={Dong, Jing and Wang, Wei and Tan, Tieniu},
  booktitle={Proceedings of the IEEE China Summit and International Conference on Signal and Information Processing (ChinaSIP)},
  year={2013}
}

@inproceedings{dang2020detectiondffd,
  title={On the detection of digital face manipulation},
  author={Dang, Hao and Liu, Feng and Stehouwer, Joel and Liu, Xiaoming and Jain, Anil K},
  booktitle={Proceedings of the IEEE/CVF Conference on Computer Vision and Pattern recognition},
  pages={5781--5790},
  year={2020}
}

@inproceedings{alvar2025divprune,
  title={Divprune: Diversity-based visual token pruning for large multimodal models},
  author={Alvar, Saeed Ranjbar and Singh, Gursimran and Akbari, Mohammad and Zhang, Yong},
  booktitle={Proceedings of the Computer Vision and Pattern Recognition Conference},
  pages={9392--9401},
  year={2025}
}

@inproceedings{shang2025llava,
  title={Llava-prumerge: Adaptive token reduction for efficient large multimodal models},
  author={Shang, Yuzhang and Cai, Mu and Xu, Bingxin and Lee, Yong Jae and Yan, Yan},
  booktitle={Proceedings of the IEEE/CVF International Conference on Computer Vision},
  pages={22857--22867},
  year={2025}
}

@article{zhang2025beyond,
  title={Beyond attention or similarity: Maximizing conditional diversity for token pruning in mllms},
  author={Zhang, Qizhe and Liu, Mengzhen and Li, Lichen and Lu, Ming and Zhang, Yuan and Pan, Junwen and She, Qi and Zhang, Shanghang},
  journal={arXiv preprint arXiv:2506.10967},
  year={2025}
}

@article{xu2025rethinking,
  title={Rethinking visual token reduction in lvlms under cross-modal misalignment},
  author={Xu, Rui and Wang, Yunke and Luo, Yong and Du, Bo},
  journal={arXiv preprint arXiv:2506.22283},
  year={2025}
}

@article{han2024filter,
  title={Filter, correlate, compress: Training-free token reduction for mllm acceleration},
  author={Han, Yuhang and Liu, Xuyang and Zhang, Zihan and Ding, Pengxiang and Chen, Junjie and Wang, Donglin and Chen, Honggang and Yan, Qingsen and Huang, Siteng},
  journal={arXiv preprint arXiv:2411.17686},
  year={2024}
}

@article{yu2026visiontrim,
  title={VisionTrim: Unified Vision Token Compression for Training-Free MLLM Acceleration},
  author={Yu, Hanxun and Li, Wentong and Qu, Xuan and Wang, Song and Chen, Junbo and Zhu, Jianke},
  journal={arXiv preprint arXiv:2601.22674},
  year={2026}
}

@article{sun2025llava,
  title={Llava-scissor: Token compression with semantic connected components for video llms},
  author={Sun, Boyuan and Zhao, Jiaxing and Wei, Xihan and Hou, Qibin},
  journal={arXiv preprint arXiv:2506.21862},
  year={2025}
}

@article{liu2025global,
  title={Global compression commander: Plug-and-play inference acceleration for high-resolution large vision-language models},
  author={Liu, Xuyang and Wang, Ziming and Chen, Junjie and Han, Yuhang and Wang, Yingyao and Yuan, Jiale and Song, Jun and Huang, Siteng and Chen, Honggang},
  journal={arXiv preprint arXiv:2501.05179},
  year={2025}
}

@article{li2025balanced,
  title={Balanced token pruning: Accelerating vision language models beyond local optimization},
  author={Li, Kaiyuan and Chen, Xiaoyue and Gao, Chen and Li, Yong and Chen, Xinlei},
  journal={arXiv preprint arXiv:2505.22038},
  year={2025}
}

@inproceedings{huang2025sida,
  title={Sida: Social media image deepfake detection, localization and explanation with large multimodal model},
  author={Huang, Zhenglin and Hu, Jinwei and Li, Xiangtai and He, Yiwei and Zhao, Xingyu and Peng, Bei and Wu, Baoyuan and Huang, Xiaowei and Cheng, Guangliang},
  booktitle={Proceedings of the Computer Vision and Pattern Recognition Conference},
  pages={28831--28841},
  year={2025}
}

@inproceedings{guo2025rethinking,
  title={Rethinking vision-language model in face forensics: Multi-modal interpretable forged face detector},
  author={Guo, Xiao and Song, Xiufeng and Zhang, Yue and Liu, Xiaohong and Liu, Xiaoming},
  booktitle={Proceedings of the Computer Vision and Pattern Recognition Conference},
  pages={105--116},
  year={2025}
}

@inproceedings{zhou2025aigi,
  title={Aigi-holmes: Towards explainable and generalizable ai-generated image detection via multimodal large language models},
  author={Zhou, Ziyin and Luo, Yunpeng and Wu, Yuanchen and Sun, Ke and Ji, Jiayi and Yan, Ke and Ding, Shouhong and Sun, Xiaoshuai and Wu, Yunsheng and Ji, Rongrong},
  booktitle={Proceedings of the IEEE/CVF International Conference on Computer Vision},
  pages={18746--18758},
  year={2025}
}

@article{guo2025language,
  title={Language-guided hierarchical fine-grained image forgery detection and localization},
  author={Guo, Xiao and Liu, Xiaohong and Masi, Iacopo and Liu, Xiaoming},
  journal={International Journal of Computer Vision},
  volume={133},
  number={5},
  pages={2670--2691},
  year={2025},
  publisher={Springer}
}

@article{wen2025spot,
  title={Spot the fake: Large multimodal model-based synthetic image detection with artifact explanation},
  author={Wen, Siwei and Ye, Junyan and Feng, Peilin and Kang, Hengrui and Wen, Zichen and Chen, Yize and Wu, Jiang and Wu, Wenjun and He, Conghui and Li, Weijia},
  journal={arXiv preprint arXiv:2503.14905},
  year={2025}
}

@article{song2024learning,
  title={On learning multi-modal forgery representation for diffusion generated video detection},
  author={Song, Xiufeng and Guo, Xiao and Zhang, Jiache and Li, Qirui and Bai, Lei and Liu, Xiaoming and Zhai, Guangtao and Liu, Xiaohong},
  journal={Advances in Neural Information Processing Systems},
  volume={37},
  pages={122054--122077},
  year={2024}
}

@article{tan2025veritas,
  title={Veritas: Generalizable deepfake detection via pattern-aware reasoning},
  author={Tan, Hao and Lan, Jun and Tan, Zichang and Liu, Ajian and Song, Chuanbiao and Shi, Senyuan and Zhu, Huijia and Wang, Weiqiang and Wan, Jun and Lei, Zhen},
  journal={arXiv preprint arXiv:2508.21048},
  year={2025}
}

@article{chen2024x2,
  title={X2-dfd: A framework for explainable and extendable deepfake detection},
  author={Chen, Yize and Yan, Zhiyuan and Cheng, Guangliang and Zhao, Kangran and Lyu, Siwei and Wu, Baoyuan},
  journal={arXiv preprint arXiv:2410.06126},
  year={2024}
}

@inproceedings{kang2025legion,
  title={Legion: Learning to ground and explain for synthetic image detection},
  author={Kang, Hengrui and Wen, Siwei and Wen, Zichen and Ye, Junyan and Li, Weijia and Feng, Peilin and Zhou, Baichuan and Wang, Bin and Lin, Dahua and Zhang, Linfeng and others},
  booktitle={Proceedings of the IEEE/CVF International Conference on Computer Vision},
  pages={18937--18947},
  year={2025}
}
\end{document}